\documentclass[lettersize,journal]{IEEEtran}
\usepackage{amsmath,amsfonts}
\usepackage{algorithmic}
\usepackage{algorithm}
\usepackage{array}
\usepackage[caption=false,font=normalsize,labelfont=sf,textfont=sf]{subfig}
\usepackage{textcomp}
\usepackage{stfloats}
\usepackage{url}
\usepackage{verbatim}
\usepackage{graphicx}
\usepackage{cite}
\usepackage{graphicx}
\usepackage{cite}
\usepackage{picinpar}
\usepackage{amsmath}
\usepackage{url}
\usepackage{flushend}
\usepackage{colortbl}
\usepackage{soul}
\usepackage{multirow}
\usepackage{pifont}
\usepackage{color}
\usepackage[table,xcdraw]{xcolor}
\usepackage{xcolor}
\usepackage{alltt}
\usepackage[hidelinks]{hyperref}
\usepackage{enumerate}
\usepackage{siunitx}
\usepackage{breakurl}
\usepackage{epstopdf}
\usepackage{pbox}
\usepackage{amsmath,amsfonts}
\usepackage{algorithmic}
\usepackage{algorithm}
\usepackage{array}
\usepackage[caption=false,font=normalsize,labelfont=sf,textfont=sf]{subfig}
\usepackage{textcomp}
\usepackage{stfloats}
\usepackage{url}
\usepackage{verbatim}
\usepackage{graphicx}
\usepackage{cite}
\usepackage{breqn}
\usepackage{makecell}
\usepackage[utf8]{inputenc} 
\begin{document}

\title{Compact Optical Single-axis Joint Torque Sensor Using Redundant Photo-Reflectors and Quadratic-Programming Calibration}

\author{Hyun-Bin Kim,~\IEEEmembership{Member,~IEEE,} Byeong-Il Ham,~\IEEEmembership{Graduate Student Member,~IEEE,} and Kyung-Soo Kim,~\IEEEmembership{Member,~IEEE}

	\thanks{
	  Manuscript created September, 2025; This research was developed by the MSC (Mechatronics, Systems and Control) lab in the KAIST(Korea Advanced Institute of Science and Technology which is in the Daehak-Ro 291, Daejeon, South Korea(e-mail: youfree22@kaist.ac.kr; byeongil\_ham@kaist.ac.kr; kyungsookim@kaist.ac.kr). (Corresponding author: Kyung-Soo Kim).  
	}
}

\markboth{}%
{Shell \MakeLowercase{\textit{et al.}}: A Sample Article Using IEEEtran.cls for IEEE Journals}


\maketitle

\begin{abstract}
This study proposes a non-contact photo-reflector-based joint torque sensor for precise joint-level torque control and safe physical interaction. Current-sensor-based torque estimation in many collaborative robots suffers from poor low-torque accuracy due to gearbox stiction/friction and current--torque nonlinearity, especially near static conditions. The proposed sensor optically measures micro-deformation of an elastic structure and employs a redundant array of photo-reflectors arranged in four directions to improve sensitivity and signal-to-noise ratio. We further present a quadratic-programming-based calibration method that exploits redundancy to suppress noise and enhance resolution compared to least-squares calibration. The sensor is implemented in a compact form factor (96~mm diameter, 12~mm thickness). Experiments demonstrate a maximum error of $0.083\%$FS and an RMS error of $0.0266$~N$\cdot$m for $z$-axis torque measurement. Calibration tests show that the proposed calibration achieves a $3\sigma$ resolution of $0.0224$~N$\cdot$m at 1~kHz without filtering, corresponding to a $2.14\times$ improvement over the least-squares baseline. Temperature-chamber characterization and rational-fitting-based compensation mitigate zero drift induced by MCU self-heating and motor heat. Motor-level validation via torque control and admittance control confirms improved low-torque tracking and disturbance robustness relative to current-sensor-based control.

\end{abstract}


\section{Introduction}


\IEEEPARstart{R}{ecent} advances in artificial intelligence and actuation technologies have accelerated research across a wide range of robotic platforms, including collaborative robots~\cite{Zhou2025troarm,Zhao2024tmecharm}, quadruped robots~\cite{di2018dynamic,hutter2016anymal,shin2022design}, and humanoid robots~\cite{unitree_website}. To enable safe and compliant interaction in human environments, precise joint-level force/torque perception and control-i.e., joint torque sensing-is critically required. However, many commercial collaborative robots estimate joint torque from motor current measurements and perform force control based on the estimated torque. For instance, the UR5e collaborative robot provides a force range of up to 50~N and a torque range of up to 10~N$\cdot$m, yet its force and torque resolutions are reported to be approximately 3.5~N and 0.2~N$\cdot$m, respectively, which are insufficient for high-precision manipulation and delicate interaction tasks~\cite{ur_website}. In particular, manipulators with high gear ratios suffer from gear-induced stiction and friction that dominate in static or near-static regimes, making precise low-torque control even more challenging when relying solely on current-based torque estimation.

A common approach for precise interaction is to mount a six-axis force/torque sensor at the end-effector. However, end-effector sensing provides only the wrench at the tool, which imposes inherent limitations for per-joint torque decomposition, compensation of gear friction/stiction, and joint-level impedance/admittance interaction control. To address these issues, robots equipped with joint torque sensors have been commercialized; nevertheless, such sensors are typically expensive and insufficiently compact, limiting their adoption and scalability.

Most joint torque sensors rely on strain-gauge technology, which offers high resolution. However, strain-gauge-based sensors are costly and require a Wheatstone bridge and signal amplification circuitry, increasing hardware complexity~\cite{nguyen2021design,fu2022design,mohandas2024design,tian2024ultracompact,kim2024design}. Moreover, strain gauges are commonly bonded manually, which limits scalability for mass production. Capacitive sensing approaches have also been investigated, where torque-induced deformation changes capacitance through variations in electrode spacing~\cite{Kim2018tmech,Kim2021tietorque,seokultrathinral2020,Chen2020torque,Lee2019icratorque}. Although capacitive interface chips such as the Analog Devices AD7147 enable relatively simple circuitry, they are sensitive to external electrostatic disturbances. Furthermore, their sampling rate is limited (typically up to 1~kHz), and their achievable precision is generally lower than that of strain-gauge-based sensors.

In this paper, to overcome the limitations of current-based torque estimation and the lack of a compact torque-sensing solution that can be directly integrated at the joint, we propose a novel photo-reflector-based joint torque sensor using a type of photo-coupler. The proposed sensor employs non-contact optical sensing to convert mechanical deformation into an electrical signal, and multiple sensing elements are placed redundantly (e.g., two sensors in each of four directions) to increase sensitivity. We further present an optimization-based calibration method to enhance precision in the low-torque regime, and experimentally validate improvements in low-torque regulation and interaction control (e.g., admittance control) compared to current-based torque control.

\subsection{Contributions}
The key contributions of this study are as follows:
\begin{itemize}
\item \textbf{Photo-reflector-based joint torque sensor architecture:}
We propose a non-contact torque sensing mechanism using a photo-reflector (photo-coupler) and a redundant four-direction sensor array to improve sensitivity.

\item \textbf{Compact and manufacturable system design:}
We develop a modular sensing/readout system that reduces wiring and assembly complexity and enables compact joint integration for scalable deployment.

\item \textbf{Redundancy-exploiting optimization-based calibration:}
We present an optimization-based calibration that leverages redundant measurements to mitigate mismatch, nonlinearity, and cross-axis coupling, improving resolution and reducing error.

\item \textbf{Experimental validation and robotic demonstration:}
We evaluate the sensor under static to dynamic conditions (resolution, nonlinearity, hysteresis) and demonstrate motor-level applicability for force/torque control.



\end{itemize}

\section{Design of the Proposed Sensor}
\subsection{Principle of the Proposed Sensor}

\begin{figure}[!t]\centering
	\includegraphics[width=0.8\columnwidth]{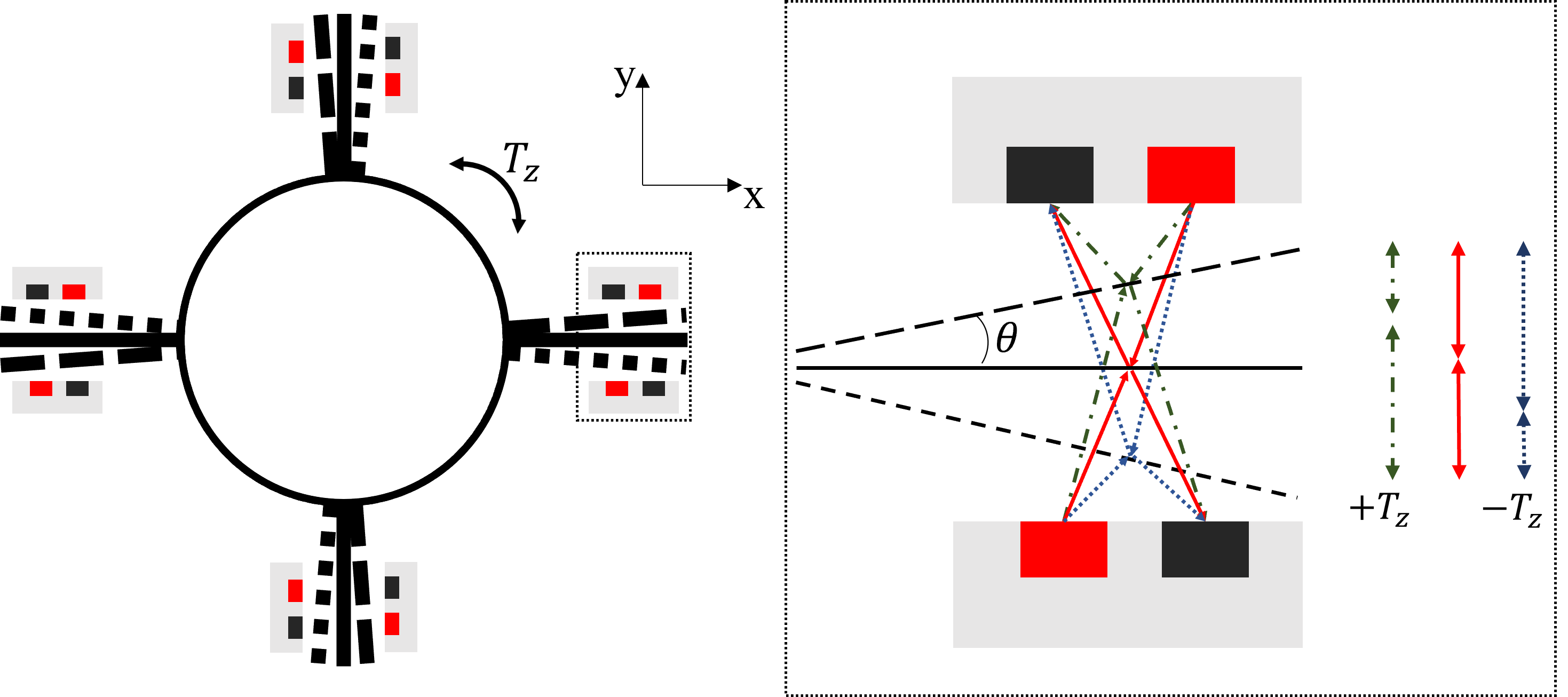}
	\caption{Configuration and operating principle of the proposed sensor. When an external torque is applied, the reflective surface rotates, altering the amount of light reflected toward the photo-reflector; this change in reflected intensity causes the output voltage to increase or decrease accordingly.}\label{principle}
\end{figure}

\begin{figure}[!t]\centering
	\includegraphics[width=0.8\columnwidth]{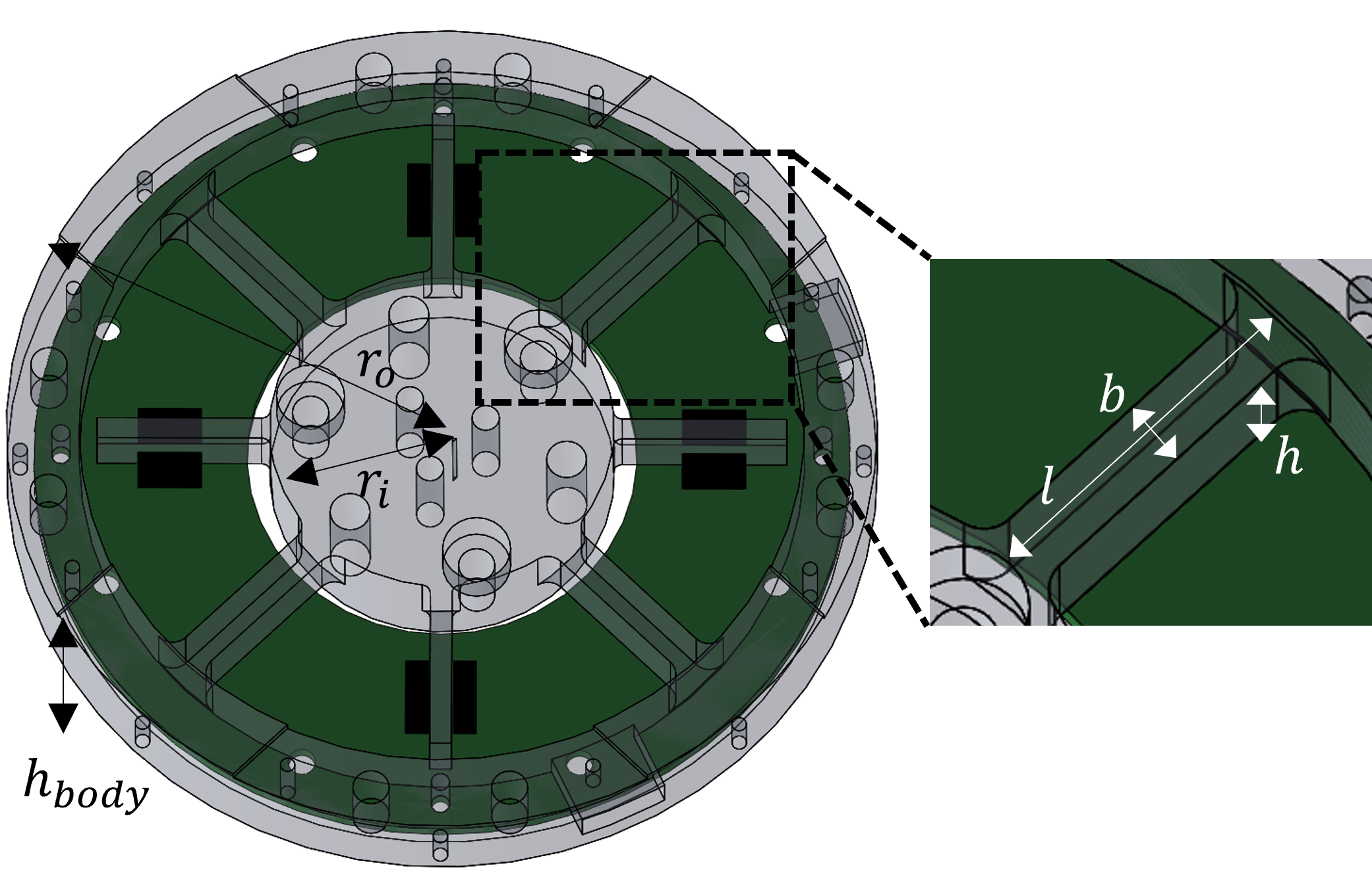}
	\caption{ Assembly of the printed circuit board (PCB) and the elastic structure, and parameters of the elastic beam.}\label{principle1}
\end{figure}
\begin{figure}[!t]\centering
	\includegraphics[width=0.8\columnwidth]{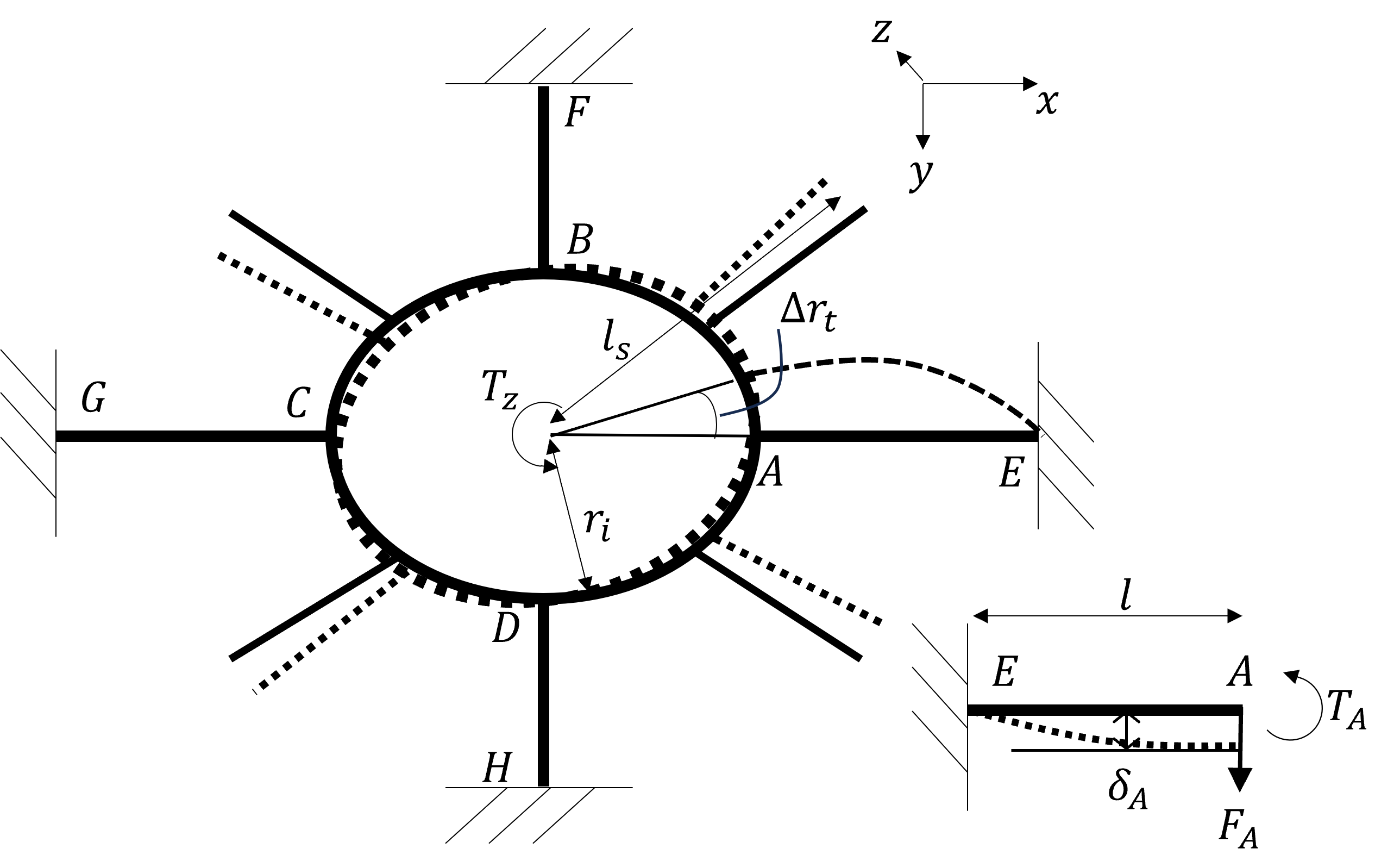}
	\caption{Schematic illustration of the elastic structure. Here, $\Delta r_t$ denotes the bending angle induced by $T_z$, and $F_A$ represents the force applied due to $T_z$.}\label{principle2}
\end{figure}

Fig.~\ref{principle} shows the overall configuration of the torque sensor. The components enclosed by the red and black boxes indicate the infrared LED and the phototransistor, respectively, and the sensor is oriented toward the horizontal plane (the $x$--$y$ plane). When a displacement occurs on the reflective surface due to a torque about the $z$-axis, the photo-reflector detects this displacement, enabling the measurement of torque. In Fig.~\ref{principle}, the green line corresponds to the case where a positive $z$-axis torque is applied, whereas the blue line corresponds to the case where a negative torque is applied. The resulting angular deflection can be expressed as $\theta$.



Fig.~\ref{principle1} illustrates the assembled configuration of the sensor, consisting of the printed circuit board (PCB) and the elastic structure, where the black region indicates the location of the photo-reflector. Here, $h_{\mathrm{body}}$ denotes the overall height of the sensor assembly, while $l$, $b$, and $h$ represent the length, width, and height of the elastic beam, respectively. By tuning these three parameters, the sensor sensitivity and measurement range can be determined. In this work, the Vishay TCRT1000 was used as the photo-reflector. $r_i$ denotes the inner radius, and $r_o$ denotes the overall radius of the sensor.

\subsection{Modeling of Elastic Beam}

As shown in Fig.~\ref{principle1}, the elastic beams (spokes) consist of four members. Although three spokes can be sufficient, this study employs four spokes to improve the resolution by exploiting redundancy. Fig.~\ref{principle2} presents a schematic illustration of the elastic structure. When a torque $T_z$ is applied, a force $F_A$ is generated at point $A$, and the resulting bending can be modeled accordingly, where $\delta_A$ denotes the displacement at $A$ under the applied force $F_A$. In addition, the angular rotation of the central part of the elastic structure can be represented by $\Delta r_t$. The radius of the central ring is denoted by $r_i$, and $l_s$ indicates the distance from the center to the sensor.

\begin{equation}
    \delta_A = \Delta r_t
\end{equation}
\begin{equation}
    \delta_A = \frac{F_A l^3 }{3EI}+\frac{F_A l}{\kappa GA}
\end{equation}
Here, $E$ denotes Young's modulus, $G$ denotes the shear modulus, and $\kappa$ denotes the shear correction factor, which is commonly taken as $5/6$ for a rectangular cross-section. The cross-sectional area is given by $A = bh$, and the second moment of area with respect to bending is $I = \frac{hb^{3}}{12}$. Therefore, the equivalent linear stiffness of a single spoke, $k_b$, can be expressed as follows~\cite{pu2021modeling,wu2023extending,ling2023dynamic}.
\begin{equation}
    k_b\equiv \frac{F_A}{\delta_A}=( \frac{l^3}{3EI}+\frac{l}{\kappa GA})^{-1}
\end{equation}
Since there are four spokes in total, the overall equivalent linear stiffness is given by
\begin{equation}
    T_z = 4 F_A r_i
\end{equation}
\begin{equation}
    F_A= k_b \delta_A = k_b r_t \theta
\end{equation}
\begin{equation}
    T_z=4(k_b r_i \theta)r_t = n k_b r_i^2 \theta
\end{equation}
\begin{equation}
    K_t \equiv \frac{T_z}{\theta}= 4 r_i^2 (\frac{l^3}{3EI}+\frac{l}{\kappa GA})^{-1} 
\end{equation}
Here, $l_s$ was set to $30$~mm and $r_i$ was fixed at $19$~mm, as constrained by the mechanical structure of the target motor. The actuator used in this study was the GIM8108 motor (SteadyWin), which can produce a maximum torque of $18$~N$\cdot$m, and accepts current-control inputs up to $2$~kHz. Conservatively, the nominal measurement target was set to $20$~N$\cdot$m, and the maximum load was defined as $400\%$. Accordingly, the design maximum torque was set to $T_z^{\max}=80$~N$\cdot$m. The corresponding design parameters are summarized in Table~\ref{parameter}.

The design was carried out using the optimization framework developed in our previous study~\cite{kim2025parameter, kim2025parameter1}. In this work, the optimization was formulated primarily based on the sensing range for the $z$-axis torque and the dimensional constraints. The resulting design was then validated via finite element analysis (FEA) in SolidWorks, as shown in Fig.~\ref{assembly}(c).

\begin{table}[!htb]
    \centering
    \caption{Value of Parameters}
    \begin{tabular}{cccccc}
    \hline\hline 
    Parameter&$l$&$b$&$h$&$r_i$&$l_s$\\\hline 
    Value (mm)&21&4&7&19&30\\
    \hline\hline 
    \end{tabular}
    
    \label{parameter}
\end{table}


In addition, since the motor diameter is $96$~mm, the sensor diameter was also set to $96$~mm. The overall dimensions are $96$~mm in diameter and $12$~mm in thickness, resulting in a slim form factor, and the total mass is $114.5$~g. The load capacity, as summarized in Table~\ref{range}, is $\pm 7700$~N for the $x$- and $y$-axis forces, $\pm 3180$~N for the $z$-axis force, $\pm 62$~N$\cdot$m for the $x$- and $y$-axis torques, and $\pm 80$~N$\cdot$m for the $z$-axis torque.

\begin{table}[!htb]
    \centering
    \caption{Specifications of the proposed torque sensor}
     \resizebox{\linewidth}{!}{
    \begin{tabular}{ccc}
    \hline\hline
        Property & Value & Unit \\ \hline
       Dimension  & $\phi$ 96 $\times$ 12 &mm \\
       Weight & 114.5 &g\\
       Load capacity & $F_{xy}$:7700,$F_{z}$:3180,$T_{xy}$:62,$T_z$:80&N,N,N$\cdot$m,N$\cdot$m\\
       Communication & CAN-FD &-\\
       Electronic bandwidth &Effective bandwidth (rise-time-limited) $\sim$ 5 ~\cite{kim2024compact} &kHz\\
       Resonant frequency& 3.407 &kHz\\
       \hline\hline
    \end{tabular}
    }
    \label{range}
\end{table}
The $\sim$5\,kHz value in Table~\ref{range} denotes the rise-time-limited effective bandwidth of the sensing chain, inferred from the measured 10--90\% rise time using $f_{\mathrm{BW}}\approx 0.35/t_{10\text{--}90}$, as characterized in our prior work~\cite{kim2024compact}.
This is distinct from the ADC sampling capability, which can be higher; the experiments/control were run at 1\,kHz for system-level synchronization. Although Table~II lists multi-axis load capacities for structural safety, this paper focuses on the $T_z$ (joint-axis) torque sensing channel and its calibration/validation; the other-axis values are reported only as mechanical load limits.
The resonant frequency of the elastic structure provides an indicative upper bound for the mechanically achievable bandwidth; in practice, the operational bandwidth is chosen sufficiently below the first structural mode.
A back-of-the-envelope estimate of the $z$-axis torsional mode was obtained as
\[
f_{n,z}=\frac{1}{2\pi}\sqrt{\frac{k_z}{J_z}},
\]
where $k_z$ is the torsional stiffness about the $z$-axis and $J_z$ is the equivalent $z$-axis inertia of the moving central part (computed from its mass distribution).
This calculation yielded $f_{n,z}\approx 5.975$~kHz.
In parallel, a SolidWorks modal analysis of the assembled elastic structure produced the first five natural frequencies of 3.407, 4.230, 5.178, 5.233, and 5.310~kHz.
Since the lowest mode (3.407~kHz) bounds all higher modes regardless of mode shape, we conservatively report 3.407~kHz as the effective structural resonance for subsequent bandwidth considerations.

\subsection{PCB Design and Assembly}
\begin{figure}
    \centering
    \includegraphics[width=0.9\linewidth]{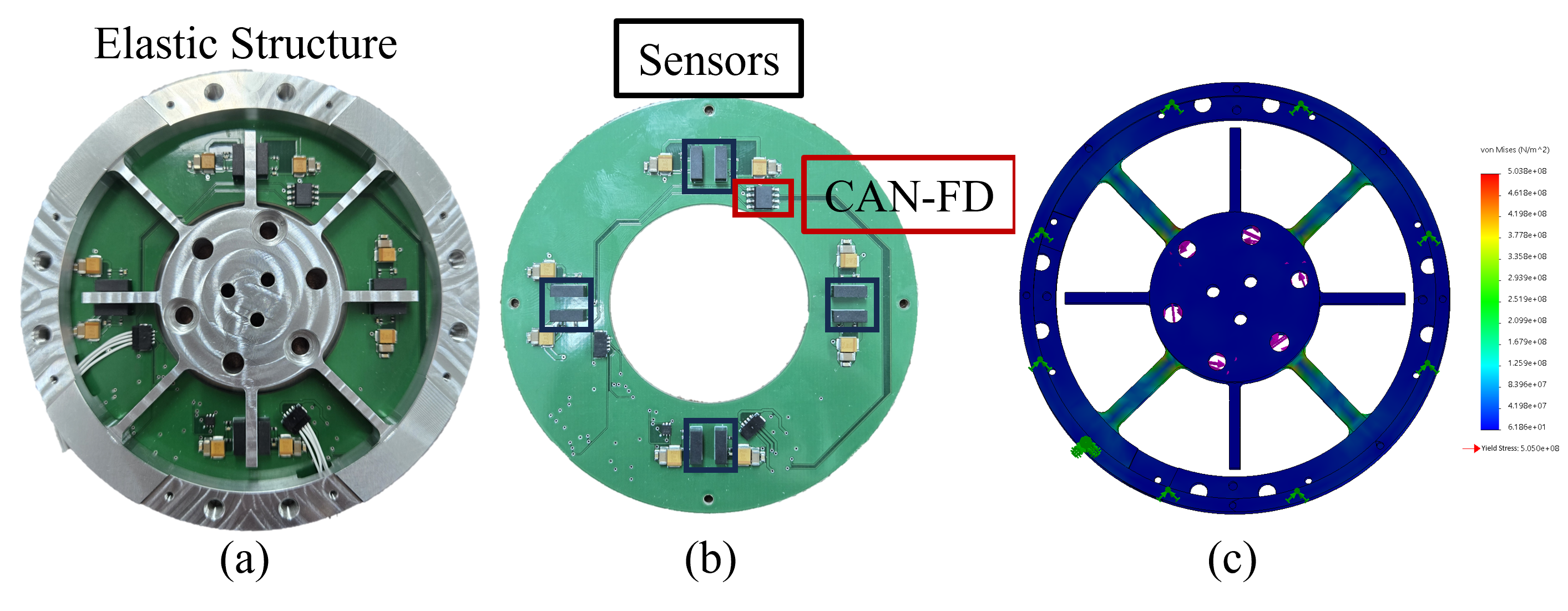}
    \caption{Assembly overview: (a) photograph of the assembled PCB and elastic structure; (b) PCB layout showing the sensor location; (c) FEA of the elastic structure.}
    \label{assembly}
\end{figure}

The complete assembly is shown in Fig.~\ref{assembly}(a). The PCB, presented in Fig.~\ref{assembly}(b), is built around an STM32H7-series micro-controller unit(MCU) and incorporates a 16-bit analog-to-digital converter(ADC). Communication is implemented via CAN-FD with a data rate of 5~Mbps. In addition, passive components were included on the PCB to stabilize the ground and improve the integrity of the analog signals, and a TI TMP117 IC was used for temperature sensing. For the photo-reflector, only one resistor is required for the LED and one resistor for the phototransistor, since the phototransistor provides inherent current amplification~\cite{kim2024compact,kim2025miniature}. Therefore, unlike conventional strain-gauge approaches, this design reduces the number of passive components and eliminates the need for an external amplifier.


\section{Calibration and Experiment}
\begin{figure}
    \centering
    \includegraphics[width=0.9\linewidth]{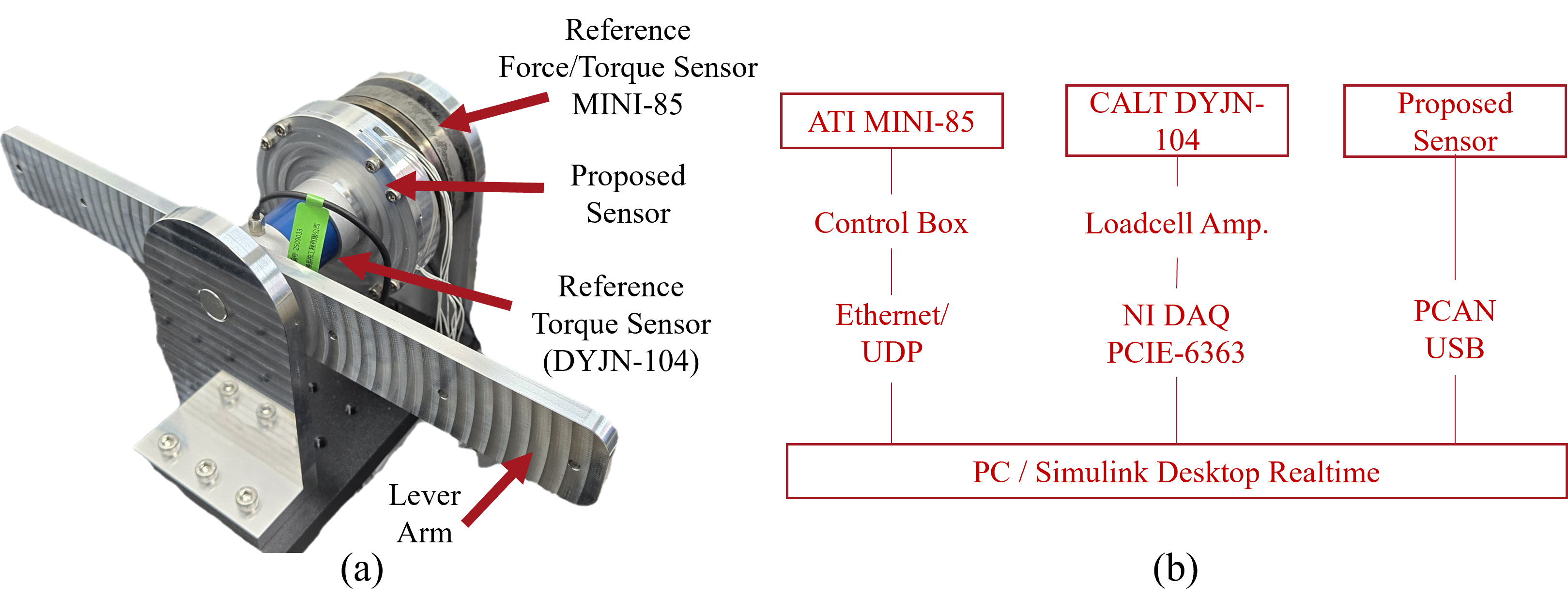}
    \caption{Calibration setup: (a) photograph of the calibration setup using the ATI MINI85, the proposed sensor, and a lever arm; (b) schematic diagram of the communication interfaces and experimental devices used for the calibration setup.}
    \label{calsetup}
\end{figure}
\subsection{Calibration Method and Experiment}
For calibration, the proposed sensor was directly coupled to a reference force/torque sensor (ATI MINI85), and experiments were conducted using a lever arm with calibrated test weights. The calibration setup is illustrated in Fig.~\ref{calsetup}(a), and the overall communication architecture and data acquisition scheme are shown in Fig.~\ref{calsetup}(b). The MINI85 data were streamed to the PC via UDP through its control box. The proposed sensor data were acquired via CAN using a PCAN-USB interface. In addition, the CALT DYJN-104 sensor was read through a load-cell amplifier and an NI DAQ (PCIe-6363) to the PC. All experiments were executed in Simulink Desktop Real-Time with a sampling rate of 1~kHz.


The specifications of each sensor are as follows. The ATI MINI85 (SI-1900-80) provides a $z$-axis torque resolution of $7/748$~N$\cdot$m and a measurement range of $80$~N$\cdot$m. It has a resonant frequency of up to $2.4$~kHz and an accuracy of $1\%$ full scale. The CALT DYJN-104 has a sensitivity of $2.0$~mV/V, a nonlinearity of $0.05\%$ full scale, and a repeatability of $0.05\%$ full scale, with a measurement range of $50$~N$\cdot$m. In our experiments, the MINI85 was used as the primary reference sensor due to its higher resolution, while the DYJN-104 was used for accuracy cross-checking and compensation.

\begin{figure}[!tb]
    \centering
    \includegraphics[width=0.8\linewidth]{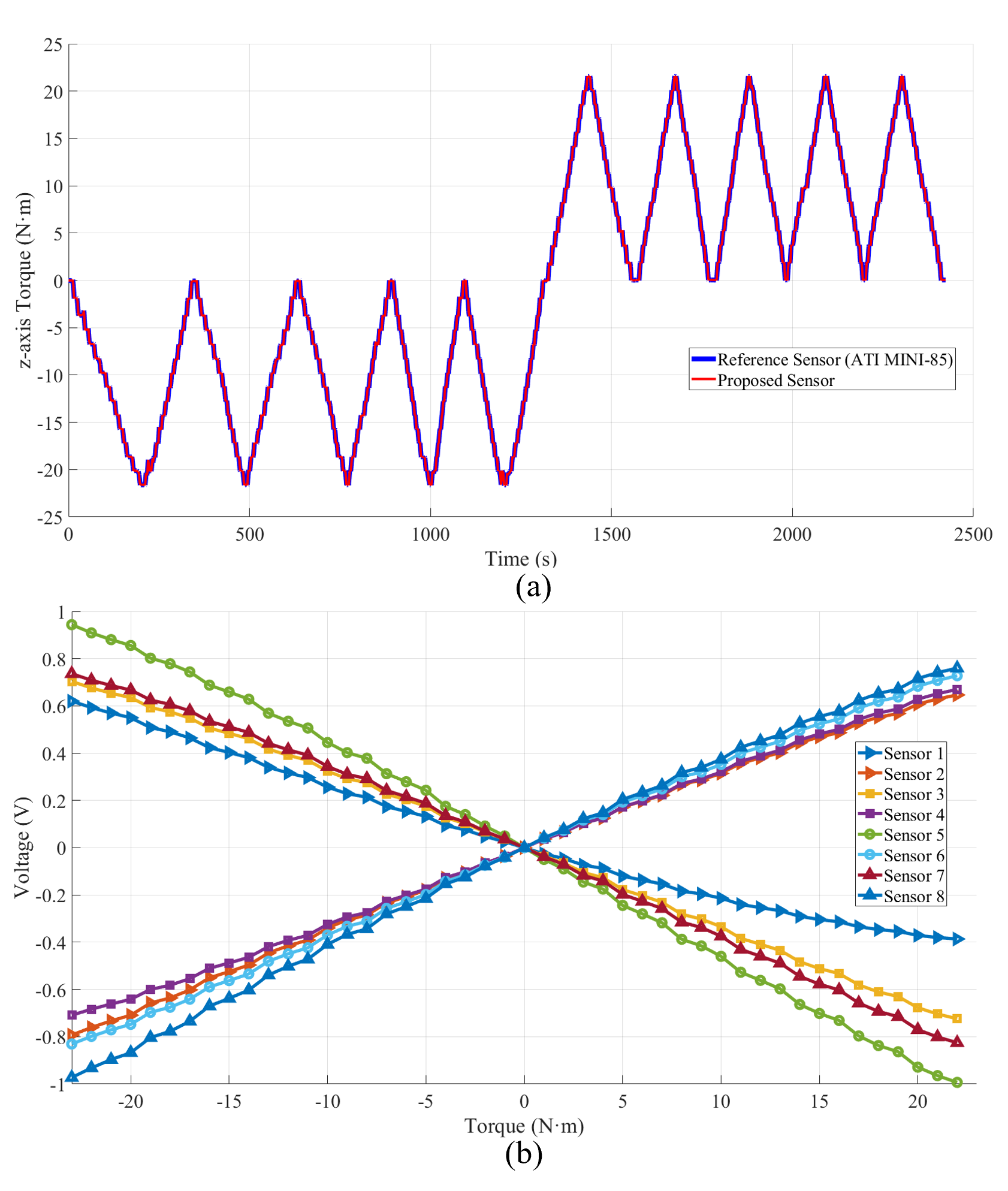}
    \caption{Calibration experiment: (a) measured responses obtained by applying test weights five times in the negative and positive directions; (b) ADC outputs of each sensor as a function of the applied torque.}
    \label{caltest}
\end{figure}

The calibration was performed using an optimization-based approach rather than the commonly used least-squares method. In the proposed sensor, the gain of each sensing element differs slightly, and the voltage response of each sensor with respect to the applied torque can be observed as shown in Fig.~\ref{caltest}(b). In this study, a total of eight sensors were employed; each sensor exhibits different gain characteristics and measured noise levels. Although a single sensing element is theoretically sufficient to measure the $z$-axis torque, using multiple sensors is essential to reduce crosstalk. Moreover, since the sensors are arranged on a single plane, redundancy arises when three or more sensors are used. By exploiting this redundancy and formulating the calibration as an optimization problem, the measurement noise can be reduced. The details are described as follows.


\begin{figure}[!htb]
    \centering
    \includegraphics[width=0.8\linewidth]{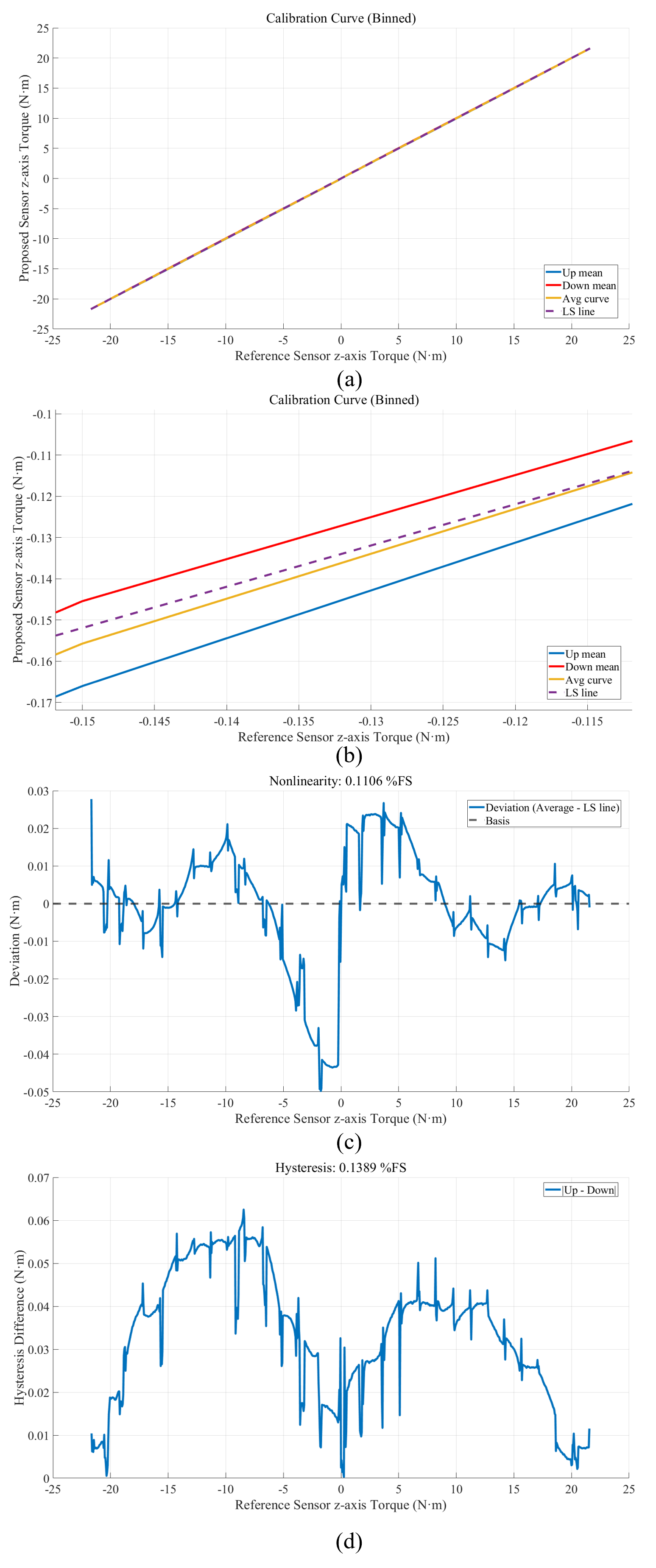}
    \caption{Nonlinearity and hysteresis experiments: (a) upper- and lower-bound average curve and the corresponding linear fit used for nonlinearity evaluation; (b) magnified view of (a); (c) nonlinearity deviation as a function of the applied torque; (d) hysteresis deviation as a function of the applied torque.}
    \label{nonlinearity}
\end{figure}
The conventional calibration method based on least squares is given in Eqs.~\ref{10} and \ref{11}. Here, $y$ denotes the ground-truth torque about the $z$-axis, and $A$ denotes the design matrix constructed from $x=[V_1\ V_2\ V_3\ V_4\ V_5\ V_6\ V_7\ V_8]$ ($V_k$ denotes the voltage output of the $k$-th sensor.), where
$A = \begin{bmatrix} x^3 & x^2 & x & 1 \end{bmatrix}$ $(N\times25)$.
Moreover, $\theta$ $(25\times1)$ denotes the calibration matrix.
\begin{equation}
    \hat{y}=A\theta
    \label{10}
\end{equation}
\begin{equation}
    \min_\theta \|A\theta-y\|_2^2
    \label{11}
\end{equation}
However, in this study, the calibration matrix was obtained by incorporating a quiet-segment-based optimization so as to preferentially utilize sensing elements with lower noise. Specifically, we propose a noise-reducing calibration method formulated as a quadratic programming problem.

Let the index set of the quiet segment be denoted by $\mathcal{I}_0$, and let the corresponding design matrix over this segment be $A_0$.
\begin{equation}
    A_{0c}=A_0 -\textbf{1} \mu^\top , \mu = \frac{1}{|\ \mathcal{I}_0|}\textbf{1}^\top A_0
\end{equation}
 Then, the output variation in the quiet segment can be defined as above. In particular, the variation of the predicted output in the quiet segment is represented as $A_{0c}\theta$, and the noise-suppression penalty can be expressed as $\gamma \|A_{0c}\theta\|_2^2$. Furthermore, let the full-scale span be denoted by $FS$.
\begin{equation}
    e_{max}=0.002\cdot FS
\end{equation}
If the maximum error is set to $0.2\%$ of full scale (FS), the constraint can be written as in the above equation. Enforcing this bound for all data samples indexed by $k$, the inequality can be rearranged into the form of Eq.~\ref{15}, and its matrix representation is given in Eq.~\ref{16}. Consequently, the calibration can be formulated as the optimization problem in Eq.~\ref{17}, whose solution yields the calibration matrix $\theta$.

\begin{equation}
    |a_k^\top \theta -y_k| \leq e_{max},\ \forall k
\label{15}
\end{equation}
\begin{equation}
    -A\theta \leq -\mathbf{y} +e_{max}\mathbf{1}, A\theta \leq \mathbf{y}+e_{max}\mathbf{1}
\end{equation}
\begin{equation}
\left[
\begin{array}{c}
A\\
-A
\end{array}
\right]\theta
\le
\left[
\begin{array}{c}
\mathbf{y}+e_{\max}\mathbf{1}\\
-\mathbf{y}+e_{\max}\mathbf{1}
\end{array}
\right].
\label{16}
\end{equation}
\begin{equation}
    \min_\theta \|A\theta-\mathbf{y}\|_2^2+\gamma\|A_{0c}\theta\|_2^2 \ \ \text{s.t.}\ |A\theta - \mathbf{y}|\leq e_{max}
    \label{17}
\end{equation}


\begin{figure*}[!htb]
    \centering
    \includegraphics[width=0.9\linewidth]{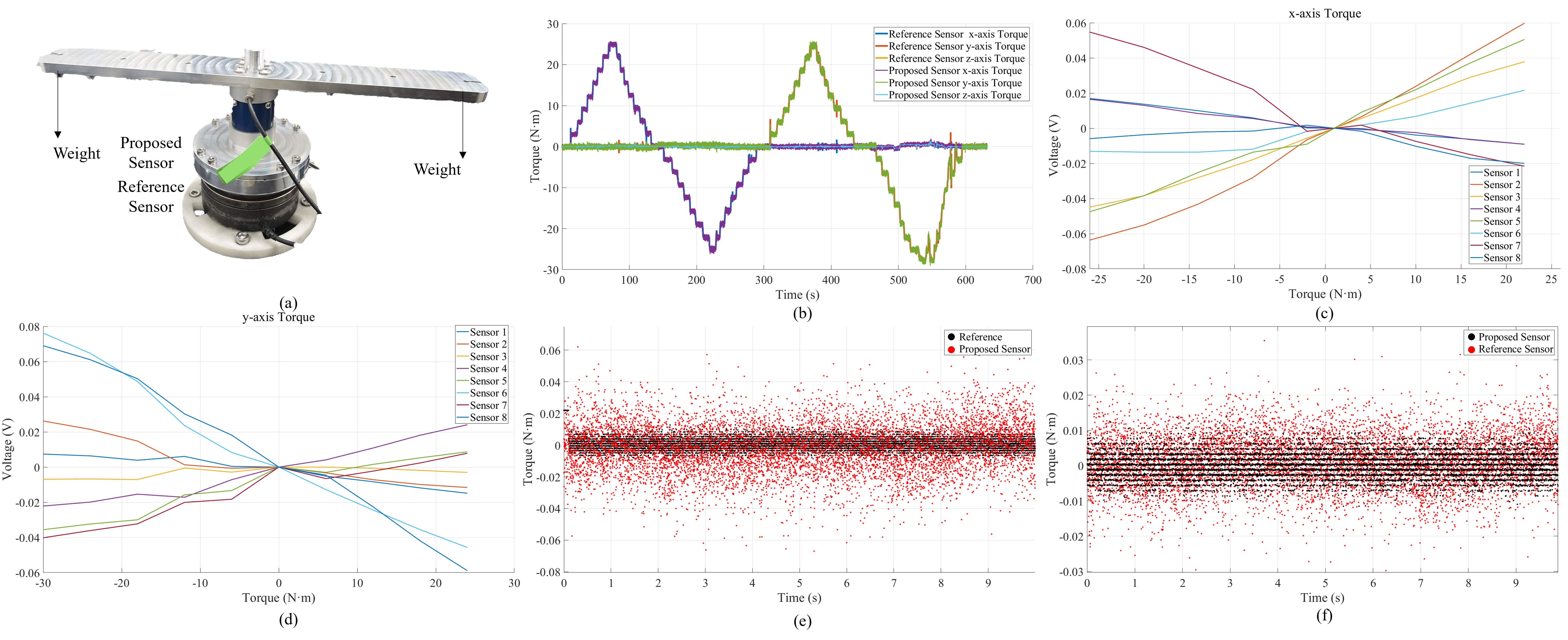}
    \caption{Crosstalk test and resolution experiments: (a) test setup for the crosstalk experiments, where weights were applied to generate moments about the $x$- and $y$-axes; (b) measured responses under the applied $x$- and $y$-axis torques; (c) sensor voltage variation under an $x$-axis torque; (d) sensor voltage variation under a $y$-axis torque; (e) noise level after calibration using the least-squares method; (f) noise level after calibration using the proposed method.}
    \label{cross}
\end{figure*}
which is a convex quadratic program (QP) since the objective is quadratic with $\mathbf{H}=2({A}^\top{A}+\gamma{A}_{0c}^\top{A}_{0c})\succeq 0$ and the constraints are linear; thus the global optimum is obtained reliably.
The quiet segment $I_0$ is selected automatically as indices satisfying near-static/no-load conditions (e.g., $|\tau|\le \tau_{\rm th}$), and ${A}_{0c}$ is built from those samples. $\tau$ denotes the reference torque (ATI MINI85).
We choose $\gamma$ from a logarithmic grid by minimizing a data-driven criterion that balances overall fit and low-torque stability, e.g., $\mathrm{RMSE}(\hat{\tau},\tau)^2+\lambda\,\mathrm{Var}(\hat{\tau})|_{I_0}$.

With this formulation, the calibration can be solved via quadratic programming. As shown in Fig.~\ref{caltest}(a), the calibration trials were conducted by applying forces to the lever arm using test weights five times in each direction (loading and unloading). The experiments were performed in MATLAB/Simulink using a real-time kernel, and the quadratic programming problem was solved using the \texttt{lsqlin} function in MATLAB.


\subsection{Sensor Evaluation}

During the calibration experiments, Fig.~\ref{caltest}(a) shows the response under an applied torque of approximately $\pm 25$~N$\cdot$m, and the corresponding outputs of each sensing element are presented in Fig.~\ref{caltest}(b). Since the target motor can produce a maximum torque of 18~N$\cdot$m, the calibration range was conservatively set to 25~N$\cdot$m. As shown in Fig.~\ref{caltest}(b), the sensor voltage changes by approximately $0.6\sim 1$~V at 25~N$\cdot$m. This corresponds to a sensitivity of approximately $0.024\sim 0.04$~V/(N$\cdot$m). With a 3.3~V-referenced ADC, this voltage range corresponds to approximately 39{,}718 digital counts.

For error analysis, Table~\ref{error} shows that the maximum percentage error was $0.083\%$ of full scale, and the Root Mean Square(RMS) error was $0.0266$~N$\cdot$m. In all error metrics reported in \%FS, the full-scale (FS) is defined as the design maximum $T^{\max}_z=80$~N$\cdot$m unless otherwise specified.

\begin{table}[!htb]
    \centering
    \caption{Error Analysis, Nonlinearity and Hysteresis}
    \resizebox{\linewidth}{!}{
    \begin{tabular}{ccccc}
    \hline\hline
         & Max Percentage Error(FS\%) & RMS Error(N$\cdot$m) &Nonlinearity(FS\%) & Hysteresis(FS\%) \\\hline
    z-axis Torque     & 0.0830&0.0266&0.111&0.139\\
    \hline
    \end{tabular}
}
    \label{error}
\end{table}

\begin{table}[!htb]
    \centering
    \caption{REPEATABILITY TEST AND CROSSTALK ANALYSIS }
    \resizebox{\linewidth}{!}{
    \begin{tabular}{ccccc}
    \hline\hline
    Repeatability (FS\%) & Crosstalk $T_x$ (FS\%) & Crosstalk $T_y$ (FS\%)\\\hline
     0.0849 & 0.148&0.126\\
    \hline
    \end{tabular}
}
    \label{repeat}
\end{table}
In addition, as summarized in Table~\ref{repeat}, the repeatability test yielded $0.0849$~FS$\%$. The nonlinearity and hysteresis were evaluated as $0.111$~FS$\%$ and $0.139$~FS$\%$, respectively, as reported in Table~\ref{error}. These results are visualized in Fig.~\ref{nonlinearity}. Fig.~\ref{nonlinearity}(a) shows the overall nonlinearity behavior, and Fig.~\ref{nonlinearity}(b) presents a magnified view around $-0.3$~N$\cdot$m. Fig.~\ref{nonlinearity}(c) and (d) plot the deviations corresponding to nonlinearity and hysteresis, respectively, showing peak values of $0.1106$~FS$\%$ and $0.1389$~FS$\%$.



A crosstalk test was also conducted. As shown in Fig.~\ref{cross}(a), the test was performed by applying moments about the $x$- and $y$-axes over a range of approximately 25~N$\cdot$m, following the procedure illustrated in Fig.~\ref{cross}(b). When the $x$- and $y$-axis torques were applied, the sensor output voltage changed by approximately 0.06~V, as shown in Fig.~\ref{cross}(c) and (d). Here, "crosstalk" refers to the apparent output of the $T_z$ channel induced by applied $T_x$/$T_y$ moments (i.e., cross-axis sensitivity), rather than multi-axis torque estimation.

The difference between the conventional least-squares (LS) calibration and the proposed quadratic-programming-based calibration can be observed in Fig.~\ref{cross}(e) and (f). With LS calibration, the standard deviation of the estimated torque was approximately 0.0160~N$\cdot$m, corresponding to a $3\sigma$-based resolution of approximately 0.0480~N$\cdot$m. In contrast, the proposed calibration reduced the standard deviation to approximately 0.00746~N$\cdot$m, yielding a $3\sigma$ resolution of approximately 0.0224~N$\cdot$m, i.e., a 2.14$\times$ improvement. This improvement is also visually evident in Fig.~\ref{cross}(f). The standard deviation was computed from 10~s of data acquired at a 1~kHz sampling rate without any filtering.

Therefore, the proposed sensor achieves a resolution of approximately 0.0224~N$\cdot$m, which is sufficient for use as a joint torque sensor. Moreover, since this value is obtained at 1~kHz without filtering, even higher resolution can be achieved by applying appropriate filtering.

\subsection{Temperature Compensation Experiment}
\begin{figure}[!htb]
    \centering
    \includegraphics[width=0.85\linewidth]{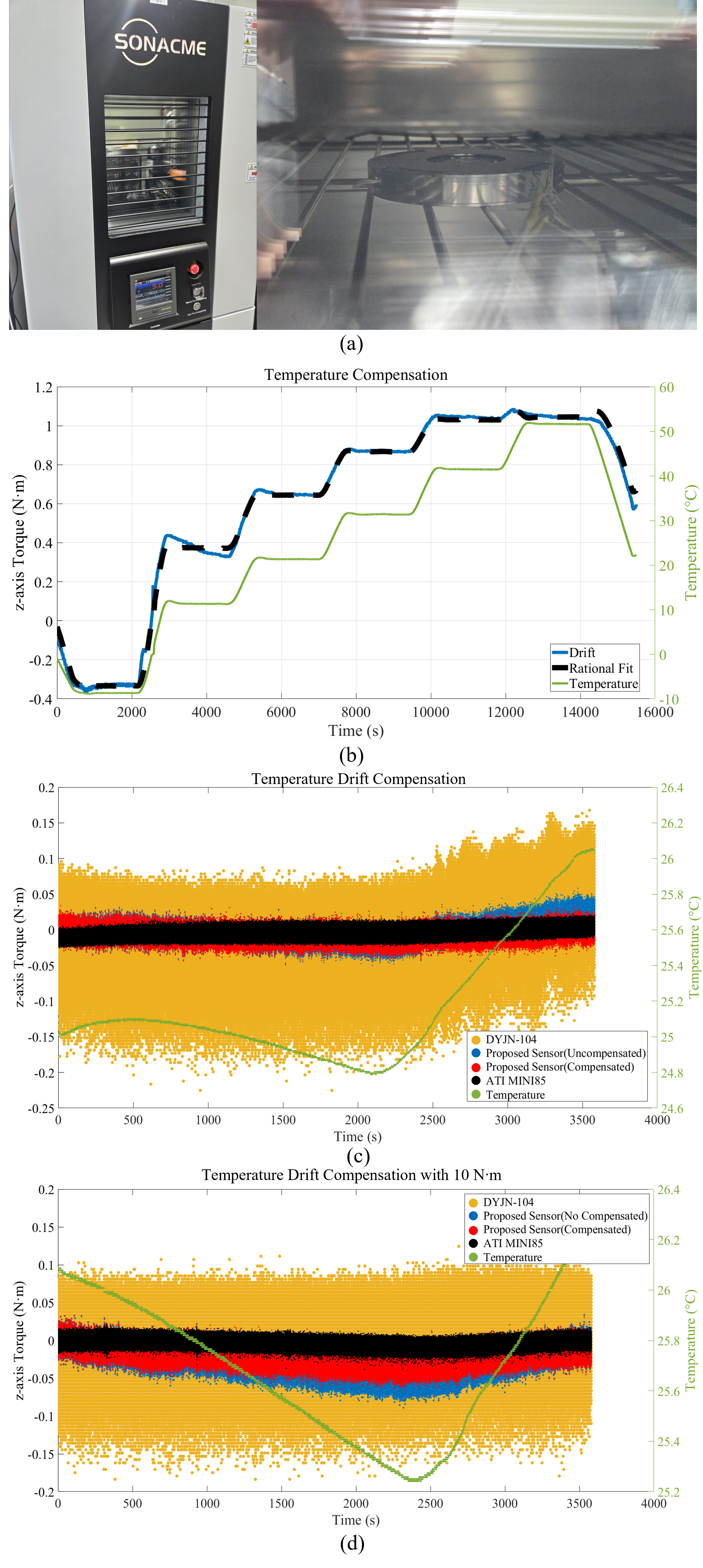}
    \caption{Temperature compensation experiment: (a) temperature chamber and the sensor placement inside the chamber; (b) temperature-dependent drift and the corresponding rational fitting curve; (c) drift at zero external torque, before and after temperature compensation; (d) drift under an external torque of 10~N$\cdot$m, before and after temperature compensation.}
    \label{Compensation}
\end{figure}
Torque sensors typically incorporate an internal MCU, which generates self-heating and induces drift; consequently, both zero drift and gain drift can occur as a function of temperature. 
Since the sensor is intended for integration with a motor, heat generated by the motor cannot be neglected; therefore, zero-drift compensation with respect to temperature is required.
To address this issue, a temperature-chamber experiment was conducted to characterize the temperature-dependent drift, as illustrated in Fig.~\ref{Compensation}. Our previous study using the same sensing principle~\cite{kim2025temp} showed that temperature-induced drift has a larger effect than gain variation. Therefore, to compensate for the zero drift, the internal temperature was measured using the on-board TMP117 IC, and the experiment was performed in a temperature chamber as shown in Fig.~\ref{Compensation}(a).

As shown in Fig.~\ref{Compensation}(b), the chamber temperature was set to $-10^\circ$C, $0^\circ$C, $10^\circ$C, $20^\circ$C, $30^\circ$C, $40^\circ$C, and $50^\circ$C. Each setpoint was held for 20~min to reach a steady state. The collected data were then used to perform rational fitting to compensate for the zero drift. The rational fitting model is given as follows.
\begin{equation}
    T_{z\ \text{drift}}=\frac{a_1 t^2+a_2t+a_3}{a_4t^2+a_5t+1}
\label{20}
\end{equation}

Here, $t$ denotes the temperature and $a_k$ denotes the $k$-th constant, where a total of five constants are used. The result of the rational fitting in Eq.~\ref{20} is shown as the black dashed line in Fig.~\ref{Compensation}(b). A five-constant model provided the best fit among the tested candidates (e.g., a fourth-order polynomial and Gaussian models), and was therefore adopted in this work. Although the fitting could alternatively be performed using a gated recurrent unit (GRU) as in~\cite{kim2025temp}, the proposed system is not as complex as a six-axis force/torque sensor, and rational fitting was found to be sufficient.

Fig.~\ref{Compensation}(c) and (d) show the results at room temperature when no external torque is applied and when a torque of 10~N$\cdot$m is applied, respectively. The black curve corresponds to the strain-gauge-based ATI MINI85. The blue markers indicate the uncompensated output, whereas the red markers represent the compensated output.



\begin{table}[!htb]
    \centering
    \caption{RMS Error of one hour drift}
    \resizebox{\linewidth}{!}{
    \begin{tabular}{ccccc}
    \hline\hline
    Sensor & MINI85& DYJN-104&Proposed Sensor&Proposed Sensor \\
    & & &(Uncompensated) &(Compensated)\\
    \hline
    Zero Torque RMS Error(N$\cdot$m)&0.0074&0.0442&0.0123&0.0091 \\
    10~N$\cdot$m Torque RMS Error(N$\cdot$m)&0.0061&0.0314&0.0326&0.0224\\
    \hline
    \end{tabular}}
    \label{tempcomp}
\end{table}

\begin{figure}[!b]
    \centering
    \includegraphics[width=\columnwidth]{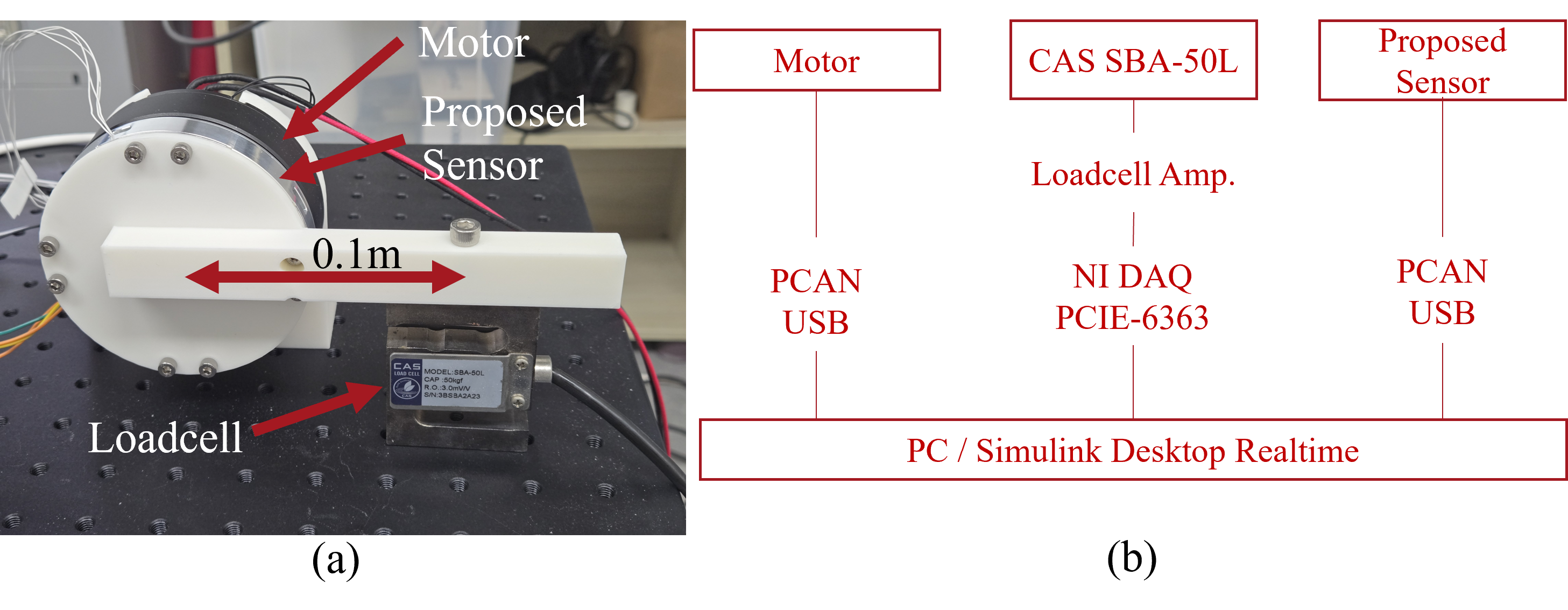}
    \caption{Motor application experiment setup: (a) photograph of the application setup integrating the motor, the proposed sensor, and the load cell; (b) schematic diagram of the communication interfaces and experimental devices used for the motor application experiments.}
    \label{motorsetup}
\end{figure}

Table~\ref{tempcomp} summarizes the RMS errors corresponding to Fig.~\ref{Compensation}(c) and (d). The RMSE of the proposed sensor was $0.0123$~N$\cdot$m and $0.0326$~N$\cdot$m before compensation, which decreased to $0.0091$~N$\cdot$m and $0.0224$~N$\cdot$m after compensation, corresponding to reductions of $27\%$ and $31\%$, respectively. In addition, a maximum temperature change of $1^\circ$C resulted in an approximately $0.014\%$ FS error variation.

\section{Motor Application Experiment}
In the motor experiments, as shown in Fig.~\ref{motorsetup}, the proposed torque sensor was mounted in front of the motor, followed by a 0.1~m lever arm and a load cell at the end of the arm to provide a redundant measurement.  The motor supports direct torque commands by using an internal current-sensing element, enabling torque-control inputs up to 2~kHz. Experiments were conducted to compare the performance obtained when using only the motor torque command versus using the proposed sensor as an additional feedback signal in the control loop.

As mentioned earlier, the motor is equipped with a gearbox, and thus exhibits backlash and stiction. Due to stiction, a minimum input torque is required to initiate motion; the stiction threshold ($\approx 0.13$~N$\cdot$m) was defined as the smallest commanded torque at which the measured motor velocity first exceeded a small threshold for a sustained duration ($>100$~ms) under chirp/PRBS/ramp excitation. 

Fig.~\ref{motorsetup}(a) shows the experimental setup, and Fig.~\ref{motorsetup}(b) illustrates the communication architecture. All experiments were performed in MATLAB/Simulink Desktop Real-Time with a 1~kHz control cycle. The load cell used in this setup was the SBA-50L (CAS), which has a measurement capacity of up to 500~N.

The experiments were conducted in three configurations to highlight the benefits of integrating the proposed sensor: (i) a step-reference tracking test, (ii) a sinusoidal tracking test, and (iii) an application-level demonstration using admittance control.

To focus on the sensor-level application in these experiments, a simple PI controller was used. For all tests, the proportional and integral gains were set to $0.12$ and $12.0$, respectively, and the anti-windup back-calculation gain was set to $5$.
\subsection{Step-Reference Tracking Test}
\begin{figure}[!htb]
    \centering
    \includegraphics[width=0.9\columnwidth]{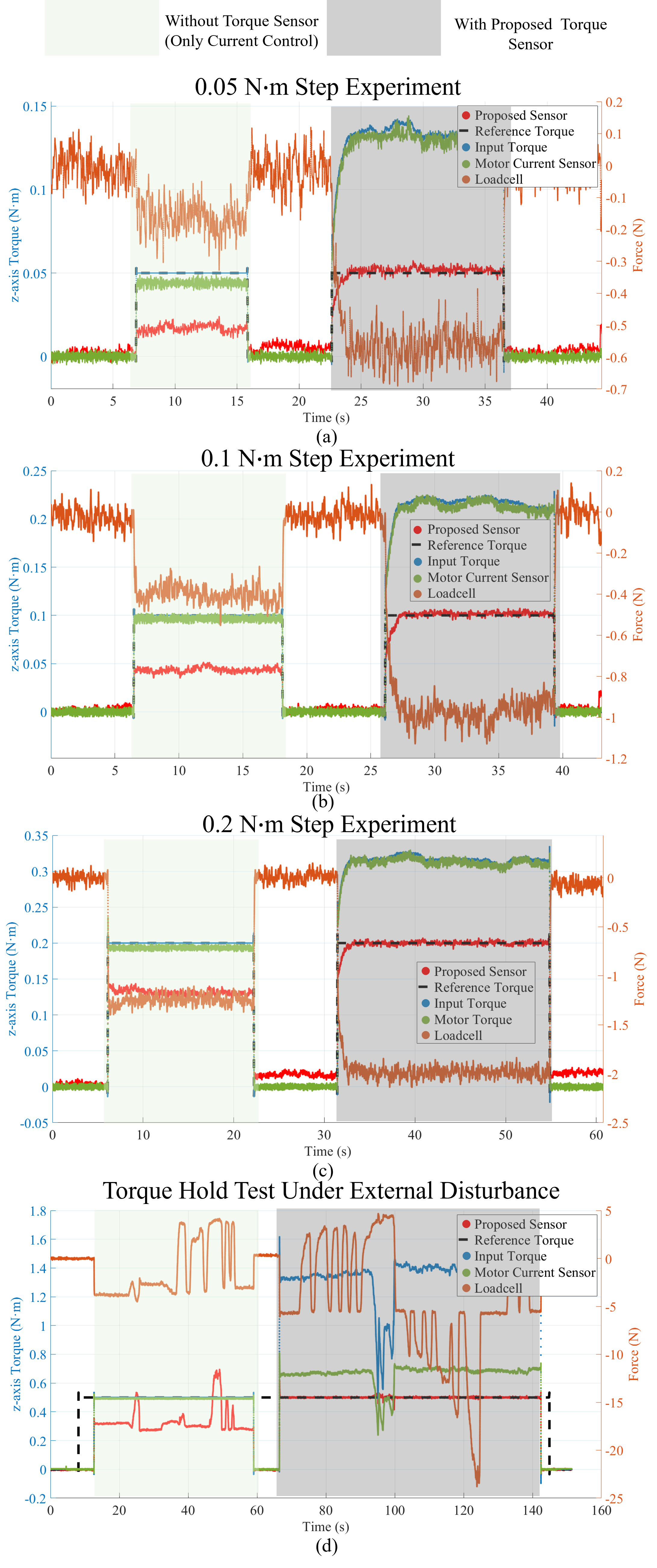}
    \caption{Step-response experiments. The light green shaded region indicates the results obtained using only the motor current sensor (without the torque sensor), whereas the gray shaded region indicates the results obtained using feedback from the proposed sensor: (a) 0.05~N$\cdot$m; (b) 0.1~N$\cdot$m; (c) 0.2~N$\cdot$m; (d) disturbance-rejection test while regulating 0.5~N$\cdot$m under an external disturbance.}
    \label{step}
\end{figure}
First, a step-reference tracking test was conducted. Step torque references of $0.05$~N$\cdot$m, $0.1$~N$\cdot$m, and $0.2$~N$\cdot$m were applied, and an additional test was performed to evaluate how well the torque could be maintained under external disturbances. Fig.~\ref{step}(a), (b), and (c) correspond to the experiments at $0.05$, $0.1$, and $0.2$~N$\cdot$m, respectively, whereas Fig.~\ref{step}(d) shows the disturbance-rejection test while regulating the torque to $0.5$~N$\cdot$m. The light green shaded region indicates torque control using the motor’s internal current sensor, while the light gray shaded region indicates torque control using the proposed sensor as feedback. For more reliable evaluation, the force measured by the load cell was also recorded, allowing the corresponding torque at the 0.1~m lever arm to be computed. Based on these measurements, the RMSE with respect to the reference torque was calculated, and the results are summarized in Table~\ref{stepre}.
\begin{table}[!htb]
    \centering
    \caption{Step-Reference Tracking Test Analysis}
     \resizebox{\linewidth}{!}{
    \begin{tabular}{ccccc}
    \hline\hline
     Experiment    &  0.05~N$\cdot$m&  0.1~N$\cdot$m&  0.2~N$\cdot$m&  0.5~N$\cdot$m with Disturbance\\ \hline
    Current Sensor& & & &\\
    RMS Error(N$\cdot$m)&  0.0329 &0.0566&0.0685 &0.1893   \\
    Torque Sensor& & & &\\
    RMS Error(N$\cdot$m)& 0.0026 &0.0026&0.0025&0.0058      \\\hline
    \end{tabular}}
    \label{stepre}
\end{table}
A key point to note is the motor stiction of approximately $0.13$~N$\cdot$m. Because the actual output torque is difficult to estimate accurately from the motor current sensor in the low-torque regime, experiments were intentionally conducted in this range. In the higher-torque regime where stiction is no longer dominant (e.g., at $0.2$~N$\cdot$m), the current-sensor-based torque control still yields insufficient output, indicating limited accuracy of the current-based torque estimate. Moreover, the fact that the output levels correspond to approximately $30\%$, $40\%$, and $65\%$ (mean values) of the expected values at the three reference levels suggests that the current-based estimate also exhibits nonlinearity. As reported in Table~\ref{stepre}, when the proposed sensor was used, the RMS error was reduced to $0.079\times$, $0.046\times$, and $0.036\times$ of the baseline values. Under external disturbances, current-sensor-only control failed to maintain regulation even at approximately $0.3$~N$\cdot$m, whereas torque-sensor feedback maintained stable regulation under disturbances exceeding $2$~N$\cdot$m.

\subsection{Sinusoidal Tracking Test}
\begin{figure*}[!htb]
    \centering
    \includegraphics[width=2\columnwidth]{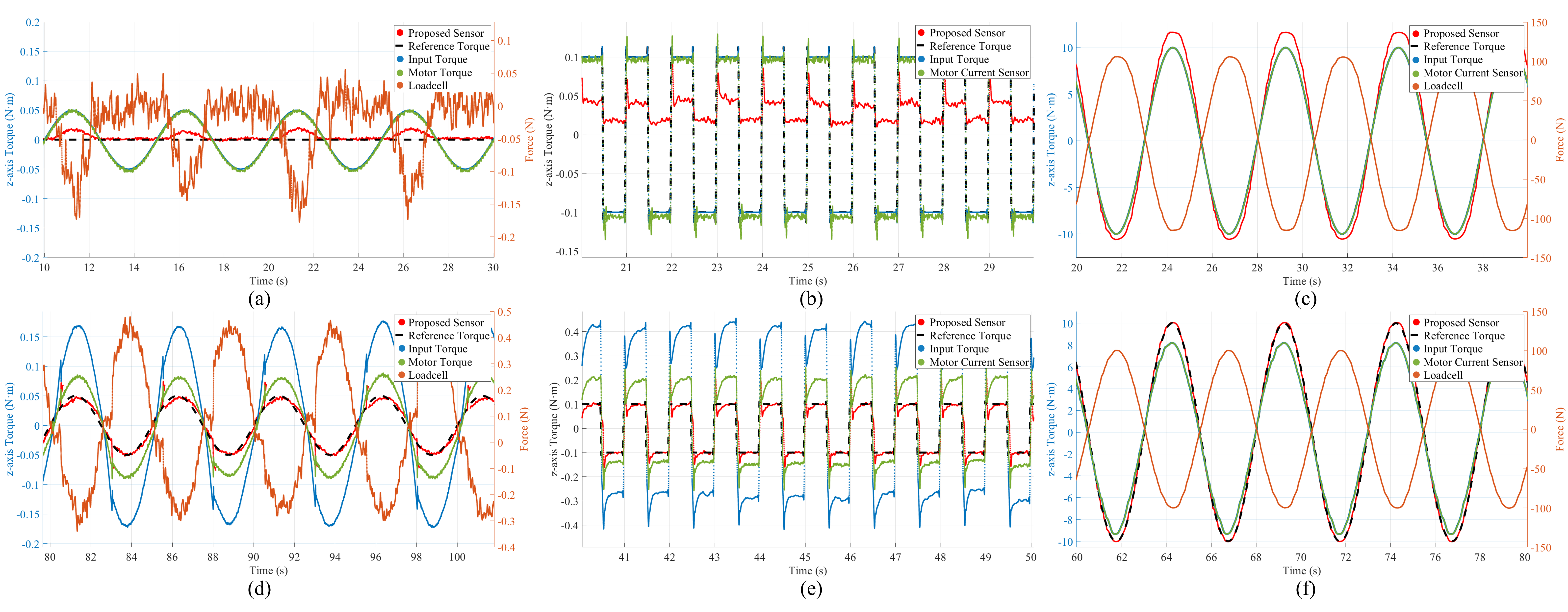}
    \caption{Sinusoidal and square-wave experiments: (a) sinusoidal test with an amplitude of 0.05~N$\cdot$m at 0.2~Hz (current-sensor-only control); (b) square-wave test with an amplitude of 0.1~N$\cdot$m at 1~Hz (current-sensor-only control); (c) sinusoidal test with an amplitude of 10~N$\cdot$m at 0.2~Hz (current-sensor-only control); (d) proposed-sensor-based control for the test in (a); (e) proposed-sensor-based control for the test in (b); (f) proposed-sensor-based control for the test in (c).}
    \label{sinwave}
\end{figure*}
Second, to evaluate whether stable control can be achieved in the low-torque regime affected by stiction, a low-amplitude sinusoidal test was performed. In addition, a low-torque square-wave input was applied to examine the behavior around the zero-crossing region, and a higher-torque test was also conducted to verify tracking performance in the non-stiction regime. Fig.~\ref{sinwave}(a--c) present the results obtained using only the motor current sensor, whereas Fig.~\ref{sinwave}(d--f) show the results when the proposed sensor is used for feedback. In the low-torque sinusoidal and square-wave tests, current-sensor-only control exhibits a one-sided actuation behavior (i.e., generating torque predominantly in a single direction), which is strongly associated with gearbox stiction. In contrast, using the proposed sensor enables accurate torque control even in the low-torque regime.
\begin{table}[!htb]
    \centering
    \caption{Sinusoidal and Square-Wave Tracking Performance}
     \resizebox{\linewidth}{!}{
    \begin{tabular}{cccc}
    \hline\hline
     Experiment    &  0.05~N$\cdot$m Sinusoidal&  0.1~N$\cdot$m Square&  10~N$\cdot$m Sinusoidal\\ \hline
    Current Sensor& & & \\
    RMS Error(N$\cdot$m)&  0.0685 &0.0924&1.2022  \\
    Peak Error(N$\cdot$m)&0.0505 &0.1426 & 2.2611\\
    Torque Sensor& & & \\
    RMS Error(N$\cdot$m)& 0.0205 &0.0436&0.1520     \\
    Peak Error(N$\cdot$m)&0.0301 &0.0797 & 0.4244\\\hline
    \end{tabular}}
    \label{sinre}
\end{table}
Table~\ref{sinre} summarizes the RMS errors for each experiment, showing that both the peak error and the RMS error are reduced for the 0.05~N$\cdot$m sinusoidal test, the 0.1~N$\cdot$m square-wave test, and the 10~N$\cdot$m sinusoidal test. Specifically, the RMS error decreases by 29.9\%, 47.2\%, and 12.6\%, respectively, while the peak error decreases by 59.6\%, 55.9\%, and 18.8\%, respectively. Note that the peak error is influenced by overshoot because torque regulation was implemented using only a PI controller; with improved controller tuning or a more advanced control scheme, a larger performance gain is expected.
\subsection{Application-Level Demonstration Using Admittance Control}
\begin{figure}[!htb]
    \centering
    \includegraphics[width=0.9\columnwidth]{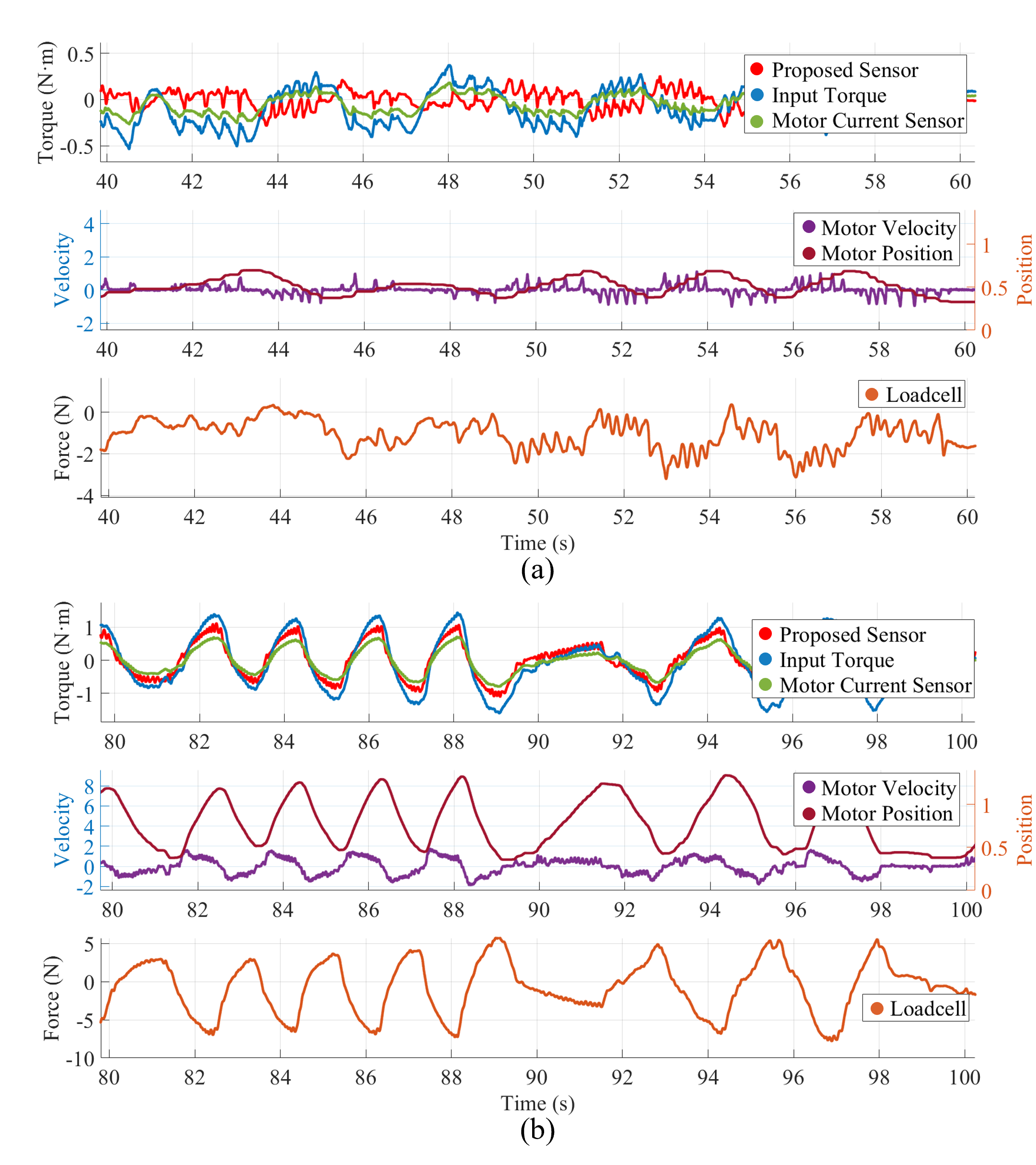}
    \caption{Admittance control experiment: (a) low-torque, low-speed interaction, exhibiting mild damping behavior; (b) high-torque, high-speed interaction, exhibiting pronounced damping behavior.}
    \label{admittance}
\end{figure}
Finally, an application-level demonstration was conducted using admittance control. A key advantage of integrating a torque sensor is that, under external interaction, the applied wrench can be estimated and used to regulate the robot joint, which is essential for interaction controllers such as admittance control~\cite{Lee2024troadmittance,Peng2024tsmcadmittance,chen2025adaptive}. The admittance controller was implemented according to Eqs.~\ref{30}-\ref{34}. The desired inertia, damping, and stiffness parameters were set to $M_d = 0.03$, $B_d = 0.6$, and $K_d = 0$, respectively, resulting in a virtual dynamics that exhibits inertia and damping without a restoring spring term. This setting is analogous to the compliant behavior commonly used in collaborative robots when a human guides the robot by hand.
\begin{equation}
    M_d \ddot{\theta}_r+B_d \dot{\theta}_r+K_d (\theta_r - \theta_0)=\tau_{ext}
    \label{30}
\end{equation}
\begin{equation}
    \ddot{\theta}_r(k)=\frac{\tau_{ext}(k)-B_d \omega_r(k)-K_d (\theta_r(k)-\theta_0)}{M_d}
    \label{31}
\end{equation}
\begin{equation}
    \omega_r(k+1)=\omega_r(k)+T_s \ddot{\theta}_r
(k)
\label{32}
\end{equation}
\begin{equation}
    e_\theta(k)= \theta_r(k)-\theta(k), \ e_\omega(k)=\omega_r(k)-\omega(k)
\label{33}
\end{equation}
\begin{equation}
    \tau _{cmd}(k)=K_p e_\theta (k) +K_v e_\omega(k)
\label{34}
\end{equation}
Here, $\theta_r$ denotes the reference angle and $\theta_0$ denotes the initial position. $\tau_{\mathrm{ext}}$ represents the external torque of the proposed sensor and $\omega_r(k)$ denotes the angular velocity at the $k$-th sample in discrete time. $T_s$ is the sampling period, which is set to $0.001$~s in this study. The terms $e_\theta$ and $e_\omega$ denote the angle error and angular-velocity error, respectively. A PD controller is then used to compute the command torque input, $\tau_{\mathrm{cmd}}$.

The experimental results are shown in Fig.~\ref{admittance}. Fig.~\ref{admittance}(a) presents the response under slowly applied external torque, whereas Fig.~\ref{admittance}(b) shows the response under rapidly applied external torque. At low speeds, the motor motion follows the direction of the applied interaction torque. At higher speeds, a resistive feel arises due to the damping term, which is also clearly reflected in the motor velocity response.

\section{Conclusion}
This paper presents a joint torque sensor based on a novel photo-reflector (photocoupler) sensing scheme and introduces an optimization-based calibration method that reduces measurement noise. By suppressing noise, the proposed approach improves the effective resolution, which is validated experimentally. The sensor is designed to be compact and self-contained, eliminating the need for additional external instrumentation compared to conventional torque-sensing solutions. Temperature-induced drift is characterized and compensated using temperature-chamber experiments, and the compensated sensor is then integrated with an actual motor to demonstrate torque-control performance as well as an application-level admittance-control experiment. These experiments provide a clear comparison between motor control performance with and without the proposed sensor, showing that precise control is achievable even in the low-torque regime where motor performance is limited.

The proposed sensor and calibration framework are expected to be beneficial for recent trends in collaborative and articulated robots, where accurate joint torque feedback is critical. In addition, this work proposes a calibration strategy that exploits redundancy in a sensor array to enhance resolution while maintaining accuracy via optimization, which is likely applicable to other sensing systems employing redundant measurements.

\bibliographystyle{IEEEtran}

	
\begin{IEEEbiography}[{\includegraphics[width=1in,height=1.25in,clip,keepaspectratio]{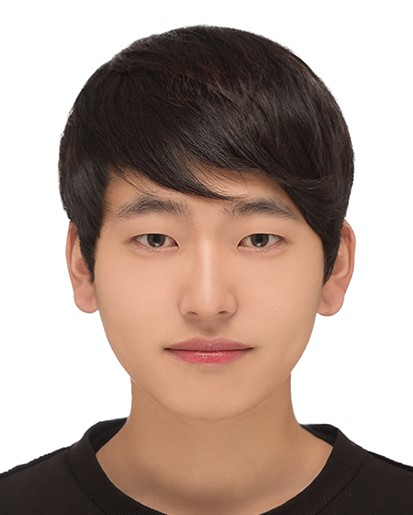}}]
{Hyun-Bin Kim}~(Member, IEEE)
~ obtained his B.S., M.S., and Ph.D. degrees in Mechanical Engineering from Korea Advanced Institute of Science and Technology(KAIST), Daejeon, Republic of Korea, in 2020, 2022, and 2025 respectively. Currently serving as a post-doctoral researcher at KAIST, his current research interests include force/torque sensors, legged robot control, robot design, and mechatronics systems.
\end{IEEEbiography}

\begin{IEEEbiography}[{\includegraphics[width=1in,height=1.25in,clip,keepaspectratio]{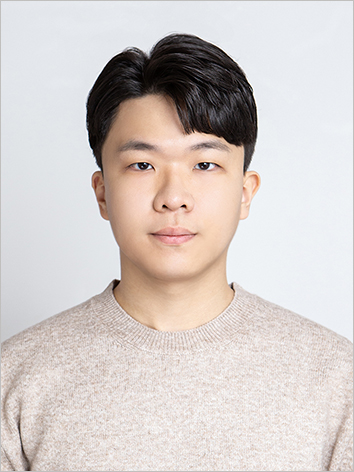}}]{Byeong-Il Ham}~(Graduate Student Member, IEEE)
~ obtained his B.S. degree in the School of Robotics from University of Kwangwoon, Seoul, and his M.S. degree in Robotics Program from Korea Advanced Institute of Science and Technology(KAIST), Daejeon, Republic of Korea, in 2022 and 2024, respectively. He is currently pursuing his Doctor Program in KAIST, Daejeon, Korea, from 2024. His current research interests include legged systems, optimal control, and motion planning.
\end{IEEEbiography}

\begin{IEEEbiography}[{\includegraphics[width=1in,height=1.25in,clip,keepaspectratio]{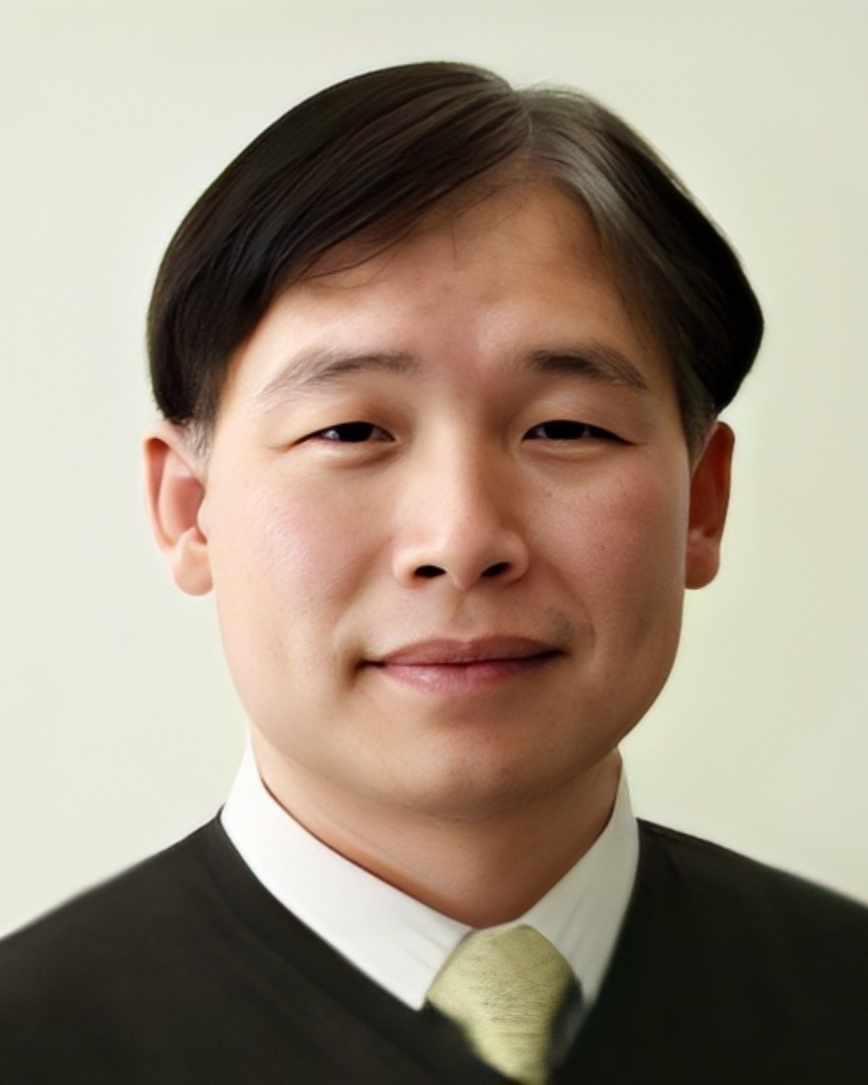}}]{Kyung-Soo Kim}~(Member, IEEE)~ obtained his B.S., M.S., and Ph.D. degrees in Mechanical Engineering from Korea Advanced Institute of Science and Technology (KAIST), Daejeon, Republic of Korea, in 1993, 1995, and 1999, respectively. Since 2007, he has been with the Department of Mechanical Engineering, KAIST. 
\end{IEEEbiography}

\end{document}